\documentclass[12pt]{article}
\usepackage{amsmath}
\usepackage{graphicx}
\usepackage{enumerate}
\usepackage{natbib}
\usepackage{url} % not crucial - just used below for the URL 
\usepackage[flushleft]{threeparttable}
\usepackage{multirow}
\usepackage{verbatim}
\newcommand{\blind}{1}

\usepackage{varwidth}
\usepackage{amsmath}
\usepackage{numprint}
\usepackage{algorithm}
\usepackage{algpseudocode}
\usepackage{multirow}
\usepackage[pdfborder={0 0 0}, colorlinks=false]{hyperref}
\usepackage{mdframed}

\algrenewcommand\algorithmicrequire{\textbf{Input:}}
\algrenewcommand\algorithmicensure{\textbf{Output:}}
\usepackage{amsmath,amssymb}

\usepackage{mathrsfs}
\usepackage{bm, xcolor}
\usepackage{mathrsfs}
\usepackage{verbatim}
\usepackage{longtable}
\usepackage{threeparttable}
\usepackage{color}
\usepackage{amsmath}

\usepackage{enumerate}
\usepackage{amssymb,bbm}
\usepackage{amsbsy}
\usepackage{amsmath}
\usepackage{amsthm}
\usepackage{amsfonts}
\usepackage{latexsym}
\usepackage{xifthen} % Allows us to put in optional arguments to questions
\usepackage{color}
\usepackage{manfnt}
\usepackage{ifthen}
\usepackage{bm}
\usepackage{caption}
\usepackage{subcaption}

\makeatletter
\def\singlespace{\def\baselinestretch{1}\@normalsize}

\newtheorem{lemma}{Lemma}%[section]
\newtheorem{theorem}{Theorem}%[section]
\newtheorem{remark}{Remark}
\@addtoreset{equation}{section}
\newtheorem{definition}{Definition}
\newtheorem{assumption}{Assumption}
\newtheorem{proposition}{Proposition}

\def\singlespace{\def\baselinestretch{1}\@normalsize}

\makeatother

\def\newpage{\vfill\eject}

\newdimen\biblioindent    \biblioindent=30pt

 at 7truept

\def\beqr{\begin{eqnarray}}
	\def\eeqr{\end{eqnarray}}
\def\beqrs{\begin{eqnarray*}}
	\def\eeqrs{\end{eqnarray*}}

\def\beq{\begin{equation}}
\def\eeq{\end{equation}}
\def\beqn{\begin{eqnarray}}
\def\eeqn{\end{eqnarray}}
\def\beqnn{\begin{eqnarray*}}
\def\eeqnn{\end{eqnarray*}}

\newcommand{\bA}{{\mathbf A}}

\newcommand{\bH}{{\mathbf H}}
\newcommand{\bI}{{\mathbf I}}

\newcommand{\bY}{{\mathbf Y}}

\newcommand{\br}{{\mathbf r}}

\newcommand{\bu}{{\mathbf u}}

\newcommand{\bw}{{\mathbf w}}

\newcommand{\bbeta}  {\boldsymbol{\beta}}

\newcommand{\btheta} {\boldsymbol{\theta}}

\theoremstyle{definition}
\makeatletter

\newcommand{\Rmnum}[1]{\expandafter\@slowromancap\romannumeral #1@}
\makeatother

\usepackage{chngcntr}
\usepackage{apptools}
\AtAppendix{\counterwithin{lemma}{section}}
\usepackage{booktabs}
\usepackage{placeins}
\ifdefined\pdfminorversion
\fi

\date{}

\begin{document}

\def\spacingset#1{\renewcommand{\baselinestretch}%
{#1}\small\normalsize} \spacingset{1}
\if1\blind
{
  \title{\bf \Large Differentially Private Statistical Inference for Rankings from Pairwise Comparisons}
  \author{
Xintao Xia\thanks{Center for Data Science, Zhejiang University}
\and
Jia Gu\footnotemark[1]\hspace{0.5em}\thanks{Corresponding author.
Email: \href{mailto:gujia@zju.edu.cn}{gujia@zju.edu.cn}.}
}
  \maketitle
} \fi

\if0\blind
{
  \bigskip
  \bigskip
  \bigskip
  \begin{center}
    \title{\bf \Large Differentially Private Statistical Inference for Rankings from Pairwise Comparisons}
\end{center}
  \medskip
} \fi

\bigskip

\begin{abstract}
Reliable inference for rankings of large language models based on human pairwise comparisons must preserve the privacy of individual preference records. To meet this need, we develop methods for score and rank inference under the Bradley–Terry model with edge differential privacy. For an objective-perturbed estimator, a new leave-one-out analysis establishes finite-sample bounds on Euclidean and entrywise estimation errors. For sufficiently dense comparison graphs, the error bounds extend to smaller privacy budgets, including regimes in which its privacy term dominates its statistical term. A uniform asymptotic expansion yields asymptotically valid confidence intervals for individual scores and pairwise score differences. To improve finite-sample performance, we account for both regularization bias and privacy-induced randomness by combining a plug-in ridge-bias correction with Gaussian–Laplace calibration. We further establish a bootstrap procedure for simultaneous rank inference, providing asymptotically valid confidence sets for individual ranks and tests of top-$K$ membership. When the comparison probability is public and known, all proposed inferential procedures require no additional privacy budget. We evaluate finite-sample performance through simulations and apply the proposed methods to rank large language models for creative writing.
\end{abstract}

\noindent%
\vfill

\newpage
\spacingset{1.9} % DON'T change the spacing!

\section{Introduction}
\label{sec:intro}

Ranking from pairwise comparison data is a fundamental statistical problem \citep{bradley1952rank}. For example, pairwise comparisons have been used to assess consumer preferences for beverages \citep{duineveld2000loglinear} and to evaluate travel packages \citep{hatzinger2007measuring}. With rapid advances in artificial intelligence (AI), ranking has become increasingly important for evaluating large language models (LLMs). As academic institutions and companies release a growing number of models, reliable rankings help identify top models, providing a basis for assessing research progress and guiding model selection for users. These rankings also have commercial implications, as companies with top-ranked AI models may attract greater investor interest \citep{aradau2026making}. Conventional benchmarks provide standardized assessments of capabilities such as knowledge and problem solving \citep{hendrycks2021measuring}. However, benchmark scores may not fully capture how users experience model responses, especially in terms of perceived empathy and emotional support \citep{zheng2023judging}. Human feedback therefore provides a basis for rankings that reflect user experience, as illustrated by Chatbot Arena, which ranks LLMs by aggregating human preferences between anonymous pairs of model responses \citep{chiang2024chatbot}. A central statistical challenge is therefore to estimate LLM rankings from human pairwise comparisons and quantify the uncertainty in these rankings.

However, releasing rankings derived from human comparisons may reveal individual preferences, even when the underlying comparison records are not released. Data protection laws, including the General Data Protection Regulation in the European Union and the California Consumer Privacy Act in the United States, further emphasize the need to protect personal information during data analysis. These concerns motivate ranking procedures with privacy guarantees while ensuring statistically valid inference. Differential privacy (DP) provides a mathematical framework for data analysis under privacy constraints \citep{dwork2006calibrating}. From a hypothesis-testing perspective, DP limits the power of any test for distinguishing neighboring datasets based on the released output \citep{wasserman2010statistical}. Achieving DP generally requires introducing randomness into the analysis \citep{dwork2014algorithmic}. The challenge in ranking inference under privacy constraints is to account for both sampling variability and privacy-induced randomness.

\subsection{Related works}
The Bradley–Terry (BT) model provides a standard framework for analyzing pairwise comparison data by associating each item with a latent parameter \citep{bradley1952rank}. These parameters determine pairwise comparison probabilities and induce a ranking. A substantial literature establishes estimation error rates and top-$K$ recovery guarantees for likelihood and spectral methods \citep{negahban2017rank,chen2019spectral,chen2022partial}. Complementing these results, \citet{han2020asymptotic,han2023general} establish consistency guarantees for maximum likelihood estimation under sparse networks. Recent developments extend the theory to general comparison graphs \citep{han2026general} and incorporate covariates within the Plackett--Luce framework \citep{dong2026statistical}. Beyond estimation, statistical inference seeks to quantify uncertainty in the estimated rank \citep{han2024inference}. \citet{gao2023uncertainty} derive asymptotic distributions for score estimators and construct confidence intervals for individual ranks, while \citet{liu2023lagrangian} develop inference for ranking properties, including pairwise ordering and top-$K$ membership. \citet{fan2025ranking} consider top-choice multiway comparison and construct simultaneous confidence intervals for score differences and individual ranks. These methods address ranking uncertainty without privacy constraints.

Estimation and inference under DP have gained attention in statistics. For example, output and objective perturbation provide general approaches to private empirical risk minimization \citep{chaudhuri2011differentially}. Asymptotic guarantees for private estimators and hypothesis tests have been established through bounded-influence methods and noisy optimization \citep{avella2021privacy,avella2023differentially}. More recently, this line of research has expanded to false discovery rate control and variable selection in high-dimensional regression \citep{cai2023differentially,gu2026learning}, as well as M-estimation by differentially private stochastic gradient descent \citep{xia2025statistical}. However, extending classical M-estimation methods to rank inference in our setting requires additional analysis to account for the identifying constraint on the latent scores, the growing dimension of the score vector, and dependence among score estimates.

More directly related to our setting, \citet{xu2025rate} develop an adaptive debiasing procedure for locally privatized pairwise comparisons and establish estimation and ranking recovery guarantees. On the inferential side, \citet{ouyang2025asymptotics} establish consistency and asymptotic normality for BT parameter estimators based on degree sequences perturbed by Laplace noise. \citet{cai2026optimal} study ranking under edge and individual differential privacy and propose a perturbed likelihood with guarantees for estimation error and top-$K$ recovery. Our work differs from that of \citet{cai2026optimal} in three respects. First, although both approaches use objective perturbation, our ridge penalty depends only on the privacy budget, whereas theirs also depends on the number of items and the comparison probability. Second, whereas their entrywise analysis tracks leave-one-out gradient descent iterates, ours directly compares the full estimator with estimators obtained by removing the comparisons involving each item. Combined with our tuning rule, this approach establishes entrywise error bounds for smaller privacy budgets when the comparison graph is sufficiently dense. Third, and most importantly, we go beyond point estimation by establishing a uniform asymptotic expansion that enables confidence intervals for individual scores and simultaneous confidence sets for ranks. To the best of our knowledge, this provides the first unified framework for inference on scores and ranks under the BT model subject to edge differential privacy.

\subsection{Main contributions}

We develop a unified framework for inference on individual scores, individual ranks, and top-$K$ membership under edge differential privacy. Our contributions are threefold.

First, we develop a new leave-one-out analysis for objective-perturbed ranking estimation and establish finite-sample bounds for both the $\ell_2$ and entrywise estimation errors. For sufficiently dense comparison graphs, our entrywise guarantee extends to smaller privacy budgets,  including regimes in which privacy-induced error dominates statistical error.

Second, we establish a uniform asymptotic expansion for the private estimator and derive asymptotically valid inference for individual scores and pairwise score differences. To improve finite-sample performance, we introduce a bias-aware procedure that combines a plug-in ridge-bias correction with Gaussian–Laplace calibration, jointly addressing regularization bias and privacy-induced randomness without additional privacy cost.

Third, we extend private score inference to simultaneous uncertainty quantification for individual ranks and top-$K$ membership. Our bootstrap procedure yields asymptotically valid rank confidence sets and top-$K$ membership tests over a potentially growing collection of items. When the comparison probability is public and known, these inferential procedures require no additional privacy budget beyond the initial private release.

\vspace{0.5cm}
\noindent\textit{Organization}. The paper is organized as follows. Section~\ref{sec:pre} introduces the BT model and edge differential privacy. Section~\ref{sec:private} presents the private estimator and procedures for score and rank inference. Section~\ref{sec: theory} establishes privacy and estimation guarantees, a uniform asymptotic expansion, and the asymptotic validity of the inference procedures. Sections~\ref{sec:simulation} and~\ref{sec:real_data} present simulation studies and a real data analysis, respectively. Section~\ref{sec:discussions} concludes with a discussion. The supplementary material (SM) provides background on differential privacy, proofs and technical lemmas, and additional numerical results.

\noindent\textit{Notation.}
We use $\mathbf1$ and $\bI$ for the all-ones vector and identity matrix, with dimensions indicated by subscripts when needed. We use $\mathbf1\{\cdot\}$ for the indicator function. For a vector $\bu$, $\|\bu\|_2$ and $\|\bu\|_\infty$ denote its Euclidean and maximum absolute entry norms, respectively. All limits are taken as $n\to\infty$. For positive sequences $a_n$ and $b_n$, $a_n\gg b_n$ means $b_n/a_n\to0$, and $X_n=o_p(a_n)$ means $X_n/a_n\to0$ in probability.

\section{Preliminaries}
\label{sec:pre}

\subsection{Ranking from pairwise comparisons}

We consider a ranking problem with $n$ items indexed by $[n]=\{1,\ldots,n\}$. Pairwise comparisons are observed at random. Specifically, each unordered pair $\{i,j\}$, $1\le i<j\le n$, is included independently with probability $p\in(0,1]$. The observed comparisons therefore form an Erd\H{o}s--R\'enyi graph $G=([n],\mathcal E)$, where $\mathcal E$ denotes the set of observed pairs. Let $A_{ij}$ denote the sampling indicator, where $A_{ij}=1$ if $\{i,j\}\in\mathcal E$ and $A_{ij}=0$ otherwise. Set $A_{ji}=A_{ij}$ and $A_{ii}=0$, and write $\bA=(A_{ij})$. The sampling probability $p$ is public and known. For each observed pair $\{i,j\}\in\mathcal E$, we record a binary comparison outcome $Y_{ij}\in\{0,1\}$, where $Y_{ij}=1$ indicates that item $i$ is preferred to item $j$, and $Y_{ij}=0$ otherwise. We set $Y_{ji}=1-Y_{ij}$, so that $Y_{ij}+Y_{ji}=1$. Conditional on the comparison graph, the observed outcomes are independent, with $Y_{ij}\sim \mathrm{Bernoulli}(\pi_{ij})$ for $\{i,j\}\in\mathcal E$, where $\pi_{ij}\in[0,1]$ is the probability that item $i$ defeats item $j$. Terms multiplied by $A_{ij}=0$ are defined to be zero. Thus, all data-dependent sums can be restricted to observed edges.

We model pairwise preferences using the BT model \citep{bradley1952rank}, which assigns each item $i\in[n]$ a latent score $\theta_i^*\in\mathbb R$. The probability that $i$ is preferred to $j$ is
$$
\pi_{ij}=F(\theta_i^*-\theta_j^*),
\qquad
F(t)=\frac{1}{1+\exp(-t)},
$$
where $F(\cdot)$ is the standard logistic distribution function. The latent score vector $\btheta^*=(\theta_1^*,\ldots,\theta_n^*)^\top$ determines the population ranking and provides the basis for ranking inference. Since the model depends only on pairwise score differences, we center the scores for identifiability and impose the following boundedness condition.

\begin{assumption}
\label{assump:bounded_score}
The true score vector $\btheta^*=(\theta_1^*,\ldots,\theta_n^*)^\top$ satisfies $\sum_{i=1}^n\theta_i^*=0$. Moreover, there exists a constant $\kappa>0$, independent of $n$, such that $\|\btheta^*\|_\infty\le \kappa$.
\end{assumption}
\noindent Bounded scores keep the pairwise comparison probabilities away from zero and one and ensure nonvanishing likelihood curvature, as commonly assumed in ranking estimation and inference \citep{chen2022partial,gao2023uncertainty,fan2025ranking}.

\subsection{Edge differential privacy}

We study statistical inference for the latent ranking parameter and the induced ranking under $(\varepsilon,\delta)$-edge differential privacy, which we define formally in this section.

\begin{definition}[$(\varepsilon,\delta)$-differential privacy]
\label{def:dp}
A randomized algorithm $\mathcal M:\mathcal D\to\mathcal R$ satisfies $(\varepsilon,\delta)$-differential privacy if, for every measurable set $B\subseteq\mathcal R$ and every pair of adjacent datasets $D,D'\in\mathcal D$ differing in at most one record,
$$
\Pr\{\mathcal M(D)\in B\} \le e^\varepsilon \Pr\{\mathcal M(D')\in B\}+\delta,
$$
where the probability is taken over the randomness of $\mathcal M$.
\end{definition}

The specified adjacency relation determines the privacy guarantee. While classical differential privacy for tabular data typically protects a single record, our pairwise-comparison setting treats an unordered pair, together with its comparison information, as the protected unit. This leads to the notion of \textit{edge adjacency}. Let $\mathcal P_n=\{(i,j):1\le i<j\le n\}$ denote the set of unordered item pairs. For each pair $1\le i<j\le n$, define 
$$
Z_{ij}=Y_{ij}\ \text{if }A_{ij}=1,\qquad
Z_{ij}=\varnothing\ \text{if }A_{ij}=0,
$$
where $\varnothing$ denotes the absence of a comparison. We write the pairwise comparison dataset as $D=(Z_{ij})_{1\le i<j\le n}$. We next define the adjacency relation for pairwise-comparison data.

\begin{definition}[Edge adjacency]
\label{def:ea}
Two comparison datasets $D=(Z_{ij})_{1\le i<j\le n}$ and $D'=(Z'_{ij})_{1\le i<j\le n}$ are \emph{edge-adjacent} if exactly one coordinate differs. Specifically, a unique pair $(i^*,j^*)$ with $1\le i^*<j^*\le n$ satisfies $Z_{i^*j^*}\neq Z'_{i^*j^*}$, while $Z_{ij}=Z'_{ij}$ for all other pairs.
\end{definition}

Thus, edge-adjacent datasets differ by the addition, removal, or modification of a comparison for a single unordered pair of items. Differential privacy under this relation protects both whether a comparison is observed and its outcome. This privacy notion is called \emph{edge differential privacy} \citep{karwa2011private,sala2011sharing}.

\section{Private estimation and uncertainty quantification}
\label{sec:private}

In this section, we construct a private estimator of the latent scores and decompose the effects of regularization and privacy perturbation. We then develop confidence intervals for individual scores, followed by simultaneous inference for ranks and top-$K$ membership.

\subsection{Private estimation}

We first introduce a differentially private estimator of the latent score vector $\btheta^*$. The estimator is based on objective perturbation \citep{chaudhuri2011differentially,iyengar2019towards} and follows the construction of \citet{cai2026optimal}. For $\btheta\in\mathbb R^n$, define $F_{ij}(\btheta)=F(\theta_i-\theta_j)$ for $1\le i<j\le n$. The negative log-likelihood based on the observed comparisons is
$$
\mathcal L_n(\btheta)=\sum_{1\le i<j\le n}A_{ij}\big[-Y_{ij}\log F_{ij}(\btheta)-(1-Y_{ij})\log\{1-F_{ij}(\btheta)\}\big].
$$
Equivalently, since $Y_{ji}=1-Y_{ij}$, this can be written as
$$
\mathcal L_n(\btheta)=\sum_{1\le i<j\le n}A_{ij}\big[-Y_{ij}\log F_{ij}(\btheta)-Y_{ji}\log\{1-F_{ij}(\btheta)\}\big].
$$
To privatize the estimator, we add a random linear perturbation to the regularized objective. Let $\bw=(w_1,\ldots,w_n)^\top$, where $w_1,\ldots,w_n \overset{\mathrm{ind}}{\sim} \mathrm{Laplace}(0,\lambda)$, independently of the comparison data, with scale parameter $\lambda > 0$. For a regularization parameter $\gamma>0$, define
\begin{equation}
\label{eq:dp_est}
\widetilde{\btheta}=\operatorname*{arg\,min}_{\substack{
\btheta\in\mathbb{R}^n}}\bigg\{\mathcal{L}_n(\btheta)+\frac{\gamma}{2}\|\btheta\|_2^2+\bw^{\top}\btheta\bigg\}.
\end{equation}
The ridge penalty ensures strong convexity of the objective, while the random linear term masks the change in the score equation caused by modifying a single comparison. These two additional terms together guarantee the privacy protection for the estimated scores. 

In non-private settings, existing likelihood-based inference procedures for ranking models \citep{gao2023uncertainty,fan2025ranking} are typically based on the unregularized maximum likelihood estimator (MLE) under the centering constraint,
$$
\widehat{\btheta}_{\mathrm{MLE}}=\operatorname*{argmin}_{\mathbf1_n^\top\btheta=0}\mathcal L_n(\btheta).
$$
The constraint respects the identification condition of the scores, since the likelihood depends only on their pairwise differences. The differentially private estimator $\widetilde{\btheta}$ is instead obtained by minimizing a regularized likelihood with an additional random linear perturbation. To understand its asymptotic behavior, we compare the estimators under the same centering normalization and discuss the effects of these two modifications separately.

\paragraph*{Effects of regularization.}
The $\ell_2$ penalty ensures strong convexity of the penalized loss function, but it also introduces shrinkage relative to the MLE. To isolate this effect, let
$$
\widehat{\btheta}_{\gamma}:=\operatorname*{arg\,min}_{\mathbf1_n^\top\btheta=0}
\bigg\{\mathcal L_n(\btheta)+\frac{\gamma}{2}\|\btheta\|_2^2\bigg\},
\qquad \gamma>0.
$$
This estimator coincides with the unconstrained ridge estimator studied for ranking estimation \citep{chen2019spectral}. To see this, shift invariance implies that $\mathbf1_n^\top\nabla\mathcal L_n(\btheta)=0$ for every $\btheta$. Summing the first-order conditions for the unconstrained ridge estimator then yields $\mathbf1_n^\top\widehat{\btheta}_{\gamma}=0$. Thus, the ridge penalty automatically enforces the centering constraint required for identifiability, making an explicit constraint unnecessary. The same shift invariance implies that the Lagrange multiplier for the constrained MLE is zero, so $\nabla\mathcal L_n(\widehat{\btheta}_{\mathrm{MLE}})=\boldsymbol0$.
By the mean-value theorem, a first-order expansion of $\nabla\mathcal L_n(\widehat{\btheta}_{\gamma})$ around $\widehat{\btheta}_{\mathrm{MLE}}$ then gives
$\nabla\mathcal L_n(\widehat{\btheta}_{\gamma})=\bH_n^\dagger(\widehat{\btheta}_{\gamma}-\widehat{\btheta}_{\mathrm{MLE}})$, where $\bH_n^\dagger$ is the integrated Hessian matrix of the form
$$
\bH_n^\dagger:=\int_0^1\nabla^2\mathcal L_n\big\{\widehat{\btheta}_{\mathrm{MLE}}+t(\widehat{\btheta}_{\gamma}-\widehat{\btheta}_{\mathrm{MLE}})\big\}\,dt .
$$
Combining this identity with the first-order condition for the penalized estimator yields
\begin{equation}
\widehat{\btheta}_{\gamma}=(\bH_n^\dagger+\gamma\bI)^{-1}\bH_n^\dagger\widehat{\btheta}_{\mathrm{MLE}} .\label{eq: ridge_displacement}
\end{equation}
To illustrate the effect on individual scores, let $h_i=(\bH_n^\dagger)_{ii}$ denote the integrated likelihood curvature for item $i$. Replacing $\bH_n^\dagger$ by its diagonal in \eqref{eq: ridge_displacement} gives the heuristic approximation
$$
\widehat\theta_{\gamma,i}
\approx
\frac{h_i}{h_i+\gamma}\widehat\theta_{\mathrm{MLE},i},
$$
which suggests stronger shrinkage for items with less comparison information. The omitted off-diagonal terms couple changes in all fitted scores, so their combined effect must be controlled to assess the inferential consequences of regularization. For the private estimator, the regularization term in \eqref{eq: heuristic expansion} accounts for this coupling through the full Hessian.

\paragraph*{Effects of perturbation.}
The linear perturbation can shift the mean of the estimated scores from zero, so the private estimator need not satisfy the centering constraint. To compare it with the penalized MLE, define the constrained perturbed penalized estimator
\begin{equation}
\label{eq:cons_penal}
\widehat{\btheta}=\operatorname*{argmin}_{\mathbf1_n^\top\btheta=0}
\bigg\{\mathcal L_n(\btheta)+\frac{\gamma}{2}\|\btheta\|_2^2+\bw^\top\btheta\bigg\}.
\end{equation}
We decompose the privacy perturbation into its average and centered components, $\bar w=\sum_{i=1}^n w_i/n$ and $\bw^\perp=\bw-\bar w\mathbf1_n$. On the centered parameter space, $\bw^\top\btheta=(\bw^\perp)^\top\btheta$. The constrained first-order condition has the form $\nabla\mathcal L_n(\widehat{\btheta})+\gamma\widehat{\btheta}+\bw=a\mathbf1_n$, where $a$ is the Lagrangian multiplier. Taking coordinate sums gives $a=\bar w$, and hence $\nabla\mathcal L_n(\widehat{\btheta})+\gamma\widehat{\btheta}+\bw^\perp=\boldsymbol0$. By strong convexity, $\widehat{\btheta}$ is therefore also the unconstrained minimizer of
$$
\mathcal L_n(\btheta)+\frac{\gamma}{2}\|\btheta\|_2^2+(\bw^\perp)^\top\btheta.
$$
The following lemma relates this centered estimator to the original private release.

\begin{lemma}[Decomposition of the privacy perturbation]
\label{lem:equiv}
Let $\widetilde{\btheta}$ and $\widehat{\btheta}$ be defined in \eqref{eq:dp_est} and \eqref{eq:cons_penal}, respectively. Then, we have
$$
\widetilde{\btheta}=\widehat{\btheta}-\frac{\bar w}{\gamma}\mathbf1_n.
$$
Consequently, $\widetilde\theta_i-\widetilde\theta_j=\widehat\theta_i-\widehat\theta_j$ for all $1\le i\ne j\le n$.
\end{lemma}

Lemma~\ref{lem:equiv} shows that the average component of the privacy noise is absorbed entirely as a uniform translation of the estimator. The centered estimator $\widehat\btheta$ can be obtained directly from the private release $\widetilde\btheta$ as $\widehat{\btheta}=\widetilde{\btheta}-\operatorname{ave}(\widetilde{\btheta})\mathbf1_n$, where $\operatorname{ave}$ denotes the average of the coordinates. Centering is a post-processing operation and preserves the privacy guarantee. It removes the common translation, while regularization and the centered perturbation remain. Inference for normalized individual scores is based on $\widehat{\btheta}$, whereas inference for pairwise score differences $\theta_i^*-\theta_j^*$ can use $\widetilde\theta_i-\widetilde\theta_j$ or, equivalently, $\widehat\theta_i-\widehat\theta_j$. These observations motivate the confidence intervals for individual scores and pairwise differences.

\subsection{Private inference for scores and pairwise differences}\label{sec: pairwise_ci}
After removing the common shift, the private estimator still differs from the non-private MLE $\widehat{\btheta}_{\mathrm{MLE}}$ because of regularization and the centered privacy perturbation. To construct private confidence intervals, we separate these effects from statistical randomness. Let
\begin{equation}
\label{eq:def_nd}
S_i=\sum_{j\ne i}A_{ij}\{Y_{ij}-F(\theta_i^*-\theta_j^*)\} \quad \mbox{ and }\quad
g_i=\sum_{j\ne i}A_{ij}F'(\theta_i^*-\theta_j^*),
\end{equation}
where $S_i$ is the sampling score and $d_i$ is the penalized curvature for item $i$ on the observed graph. Further define the penalized loss Hessian and the associated regularization term by
$$
\bH_\gamma(\btheta)=\nabla^2\mathcal L_n(\btheta)+\gamma\bI_n
\quad \mbox{ and } \quad
\bbeta_\gamma=\gamma\bH_\gamma(\btheta^*)^{-1}\btheta^*.
$$
By the proof of Theorem~\ref{thm:expansion}, the analysis underlying its expansion gives, for each fixed $i$,
\begin{equation}
\label{eq: heuristic expansion}
\widehat\theta_i-\theta_i^*
=\frac{S_i}{g_i}-\beta_{\gamma,i}-\frac{w_i^\perp}{g_i}
+o_p\{(np)^{-1/2}\}.
\end{equation}
The first term captures sampling variability and, under the stated conditions, is asymptotically equivalent to the leading fluctuation of the non-private MLE. The second and third terms represent regularization bias and centered privacy perturbation, respectively. Under the stated conditions, regularization bias and centered privacy perturbation are asymptotically negligible but can affect finite-sample coverage.

\noindent\textbf{Wald confidence intervals.}
Under regularity conditions, the regularization bias and centered privacy perturbation are asymptotically negligible relative to statistical sampling variation. The centered private estimator $\widehat\btheta$ therefore has the same first-order normal approximation as the constrained non-private MLE coordinate-wise. Such first-order equivalence motivates Wald inference for private M-estimators. Specifically, we estimate the standard deviation of the leading statistical term $S_i/g_i$ using the plug-in estimator $\widetilde\rho_i:=\widetilde g_i^{-1/2}$, where $\widetilde g_i=p\sum_{j\ne i}F'(\widetilde\theta_i-\widetilde\theta_j)$. For a prespecified pair $(i,j)$, the respective leading statistical terms are asymptotically independent, giving the pairwise standard error estimator $\widetilde\rho_{ij}:= (\widetilde\rho_i^2+\widetilde\rho_j^2)^{1/2}$. For $\alpha\in(0,1)$, let $z_{1-\alpha/2}$ denote the $(1-\alpha/2)$ standard normal quantile. The two-sided Wald intervals for a prespecified score $\theta_i^*$ and pairwise difference $\theta_i^*-\theta_j^*$ are
\begin{equation}
\label{eq:private_ci_score}
\mathrm{CI}_i(1-\alpha)=\big[
\widehat\theta_i-z_{1-\alpha/2}\widetilde\rho_i,
\widehat\theta_i+z_{1-\alpha/2}\widetilde\rho_i\big] \quad \mbox{ and }
\end{equation}
\begin{equation}
\label{eq:private_ci_compare}
\mathrm{CI}_{ij}(1-\alpha)=\big[
(\widetilde\theta_i-\widetilde\theta_j)-z_{1-\alpha/2}\widetilde\rho_{ij}, (\widetilde\theta_i-\widetilde\theta_j)+z_{1-\alpha/2}\widetilde\rho_{ij}\big],
\end{equation}
respectively. These intervals take the standard Wald form, with their centers and standard errors obtained by post-processing the private estimator. 

In finite samples, particularly when comparison information is limited, or the privacy budget $\varepsilon$ is small, regularization bias and privacy perturbation may be non-negligible relative to statistical sampling variation. Ignoring these effects may lead to undercoverage of the Wald confidence intervals. We therefore develop a bias-corrected procedure that accounts for both effects in the subsequent calibrations.

\noindent\textbf{Bias correction.}
The leading regularization bias is $-\beta_{\gamma,i}$, so we correct the center by adding a plug-in estimate of $\beta_{\gamma,i}$. To estimate $\bbeta_\gamma$, we replace each graph indicator $A_{ij}$ in $\bH_\gamma(\btheta)$ by its public sampling probability $p$ and evaluate the resulting matrix at the centered private estimator $\widehat\btheta$. Denote this matrix by $\widetilde\bH_\gamma$, whose entries are
$$
(\widetilde\bH_\gamma)_{ii}=\widetilde g_i+\gamma,
\qquad
(\widetilde\bH_\gamma)_{ij}
=-pF'(\widehat\theta_i-\widehat\theta_j),\quad i\ne j.
$$
The diagonal entries use the previously defined $\widetilde g_i$ because centering preserves score differences: $\widehat\theta_i-\widehat\theta_j=\widetilde\theta_i-\widetilde\theta_j$. We then estimate $\bbeta_\gamma$ by \(\widetilde\bbeta_\gamma=\gamma\widetilde\bH_\gamma^{-1}\widehat\btheta\), and the bias-corrected estimator is constructed as
\begin{equation}
\label{eq:debiase}
\widetilde\theta_i^{(\mathrm{bc})}
=\widehat\theta_i+\widetilde\beta_{\gamma,i},
\qquad i\in[n].
\end{equation}

\noindent\textbf{Gaussian--Laplace calibration.}
Correcting the regularization bias leaves the random privacy perturbation in the expansion \eqref{eq: heuristic expansion}. To account for privacy-induced randomness in finite samples, we use Gaussian––Laplace calibration motivated by the expansion. Although the centered noise coordinates $\{w_i^\perp\}_{i=1}^n$ are dependent and are not exactly Laplace, the contribution of the noise average $\bar w$ is asymptotically negligible on the standard-error scale for a fixed score or pair. This permits calibration using independent Laplace draws. Specifically, with the private estimator held fixed, let $Z\sim N(0,1)$ and $W,W_1,W_2\overset{\mathrm{i.i.d.}}{\sim}\operatorname{Laplace}(0,\lambda)$ be fresh, mutually independent draws, independent of the fit. For each prespecified score or pair, define its Gaussian--Laplace calibration variable on the sampling-standard-error scale:
$$
T_i^{\mathrm{mix}}=Z+\widetilde\rho_i W,
\qquad
T_{ij}^{\mathrm{mix}}
=Z+\frac{\widetilde\rho_i^2W_1+\widetilde\rho_j^2W_2}{\widetilde\rho_{ij}}.
$$
The symmetry of the Laplace distribution justifies the plus signs. Let $q_i^{\mathrm{mix}}(u)$ and $q_{ij}^{\mathrm{mix}}(u)$ denote the conditional $u$-quantiles. Since both distributions are symmetric about zero, the bias-corrected two-sided confidence intervals for $\theta_i^*$ and $\theta_i^*-\theta_j^*$ are constructed as
\begin{equation}
\label{eq:bc_ci_score}
\mathrm{CI}_i^{(\mathrm{bc})}(1-\alpha)=\left[
\widetilde\theta_i^{(\mathrm{bc})}-q_i^{\mathrm{mix}}(1-\alpha/2)\widetilde\rho_i,\widetilde\theta_i^{(\mathrm{bc})}+q_i^{\mathrm{mix}}(1-\alpha/2)\widetilde\rho_i\right]\quad \mbox{ and }
\end{equation}
\begin{equation}
\label{eq:bc_ci_pairwise}
\mathrm{CI}_{ij}^{(\mathrm{bc})}(1-\alpha)=\left[
(\widetilde\theta_i^{(\mathrm{bc})}-\widetilde\theta_j^{(\mathrm{bc})})
-q_{ij}^{\mathrm{mix}}(1-\alpha/2)\widetilde\rho_{ij}, (\widetilde\theta_i^{(\mathrm{bc})}-\widetilde\theta_j^{(\mathrm{bc})})
+q_{ij}^{\mathrm{mix}}(1-\alpha/2)\widetilde\rho_{ij}\right],
\end{equation}
respectively. Theorem~\ref{thm:plugin_inference} establishes asymptotic coverage $1-\alpha$ for both procedures.

\begin{remark}[No additional privacy cost]
A notable feature of the proposed inference procedure is that neither the variance estimation nor the bias correction requires an additional privacy budget. This contrasts with private inference for general M-estimation, where estimating covariance matrices often requires separate privatization steps \citep{avella2023differentially,xia2025statistical}. In our setting, all quantities entering the confidence intervals are functions of the private estimator $\widetilde\btheta$ and public parameters. Bias correction, plug-in variance estimation, and interval construction are post-processing operations and therefore preserve the original $(\varepsilon,0)$-edge DP guarantee.
\end{remark}

\subsection{Private inference for ranks}\label{sec:rank_inference}
Section~\ref{sec: pairwise_ci} develops confidence intervals for prespecified scores and pairwise score differences. Inference for an item's rank instead requires simultaneous control of its comparisons with the other items. We therefore use bootstrap calibration to obtain simultaneous bounds on the relevant score differences, which in turn yield two-sided confidence intervals for the rank of a prespecified item $i\in[n]$ and one-sided lower bounds for top-$K$ inference. Let
$$
r(i)=1+\sum_{k\ne i}\mathbf1\{\theta_k^*>\theta_i^*\}
$$
denote the population rank of item $i$. Smaller rank values correspond to higher scores, and this definition allows items with equal scores to share a rank. Let $\varnothing\ne\mathcal I\subseteq[n]$ be a prespecified set of items whose ranks are of interest, whose size may grow with $n$: Choosing $\mathcal I=\{i\}$ targets an individual rank, while $\mathcal I=[n]$ gives simultaneous inference for all ranks.

\noindent\textbf{Bootstrap construction.}
Consider the ordered comparison family and contrasts by 
$$
\mathcal P_{\mathcal I}:=\{(k,i):i\in\mathcal I,\ k\ne i\}, \quad  \Delta_{ki}^*:=\theta_k^*-\theta_i^*, \quad \mbox{ and } \quad \widetilde\Delta_{ki}:=\widetilde\theta_k-\widetilde\theta_i,
$$
respectively, where $(k,i)\in \mathcal P_{\mathcal I}$. We use the pairwise standard error estimator  from Section~\ref{sec: pairwise_ci}, denoted here by $\widetilde\sigma_{ki}=\widetilde\rho_{ki}=\big(\widetilde\rho_k^2+\widetilde\rho_i^2\big)^{1/2}$. To calibrate simultaneous bounds, we approximate the joint distribution of the contrast errors using the leading terms in the asymptotic expansion of the private estimator. Let $g_i:=\sum_{j\ne i}A_{ij}F'(\theta_i^*-\theta_j^*)$, which is the $i$th diagonal entry of the Hessian $\nabla^2\mathcal L_n(\btheta^*)$. With $S_i$ defined as in \eqref{eq:def_nd}, the uniform expansion in Theorem~\ref{thm:expansion} identifies $S_i/g_i$ as the leading estimation error after removal of the common shift. Since this shift cancels in pairwise differences, the expansion motivates the approximation
$$
\widetilde\Delta_{ki}-\Delta_{ki}^*
\approx \frac{S_k}{g_k}-\frac{S_i}{g_i},
\qquad (k,i)\in\mathcal P_{\mathcal I}.
$$
The simultaneous inference result of \citet{fan2025ranking} uses a Gaussian multiplier bootstrap and requires a growing number of observations per sampled comparison set. In our pairwise setting, we construct a parametric bootstrap of these leading terms that allows one outcome per sampled pair.
Specifically, let $B$ denote the number of bootstrap replications. Conditional on the private estimator $\widetilde\btheta$, the replications are independent. For each $b=1,\ldots,B$ and $1\le i<j\le n$, generate $A_{ij}^{(b)}\sim\mathrm{Bernoulli}(p)$ independently, where $p$ is public and known. Set $A_{ji}^{(b)}=A_{ij}^{(b)}$ and $A_{ii}^{(b)}=0$. Conditional on the simulated graph, generate independent outcomes by
$$
Y_{ij}^{(b)}\sim\mathrm{Bernoulli}\{F(\widetilde\theta_i-\widetilde\theta_j)\},\qquad Y_{ji}^{(b)}=1-Y_{ij}^{(b)}.
$$
For each simulated sample $(\bA^{(b)},\bY^{(b)})$, compute the likelihood scores and the diagonal entries of the negative log-likelihood Hessian at the private estimator:
$$
S_i^{(b)}=\sum_{j\ne i}A_{ij}^{(b)}\{Y_{ij}^{(b)}-F(\widetilde\theta_i-\widetilde\theta_j)\},\quad g_i^{(b)}=\sum_{j\ne i}A_{ij}^{(b)}F'(\widetilde\theta_i-\widetilde\theta_j).
$$
Set $S_i^{(b)}/g_i^{(b)}=0$ when $g_i^{(b)}=0$. Under the stated conditions, the probability that a replication has at least one isolated vertex is at most $n(1-p)^{n-1}$, so this convention has a negligible distributional effect. The ratio $S_i^{(b)}/g_i^{(b)}$ simulates the leading estimation error for item $i$ under the model evaluated at $\widetilde\btheta$. Within each replication, all contrasts are computed from the same simulated graph and outcomes, reproducing their dependence under the fitted model. The original plug-in standard errors are held fixed across replications.

With $p$ public and known, the bootstrap uses only the private estimator and independent simulation randomness, so the resulting critical values and inferential outputs incur no additional privacy cost. If $p$ is unknown, it can be estimated using a separate differentially private procedure; see, for example, \citet{ullman2019efficiently}. Incorporating such an estimate requires accounting for its privacy cost and estimation error. The same bootstrap replications are used for the two-sided and one-sided calibrations below.

\noindent\textbf{(i) Two-sided rank confidence intervals.}
For two-sided inference, define the maximum absolute standardized contrast error by
$$
T_{\mathcal I}=\max_{(k,i)\in\mathcal P_{\mathcal I}}\bigg|\frac{\widetilde\Delta_{ki}-\Delta_{ki}^*}{\widetilde\sigma_{ki}}\bigg|.
$$
Using the simulated replications above, compute the bootstrap statistic
$$
T_{\mathcal I}^{(b)}=\max_{(k,i)\in\mathcal P_{\mathcal I}}\bigg|\frac{S_k^{(b)}/g_k^{(b)}-S_i^{(b)}/g_i^{(b)}}{\widetilde\sigma_{ki}}\bigg|.
$$
The simulated critical value is the empirical $(1-\alpha)$-quantile,
$$
\widehat q_{1-\alpha}=\inf\bigg\{t:\frac1B\sum_{b=1}^B\mathbf1\{T_{\mathcal I}^{(b)}\le t\}\ge1-\alpha\bigg\}.
$$
Using the bootstrap critical value, construct the pairwise confidence intervals for each pair $(k,i)\in\mathcal P_{\mathcal I}$ with endpoints
$$
\widetilde C_L(k,i)
=\widetilde\Delta_{ki}
-\widehat q_{1-\alpha}\widetilde\sigma_{ki},\quad 
\widetilde C_U(k,i)
=\widetilde\Delta_{ki}
+\widehat q_{1-\alpha}\widetilde\sigma_{ki}.
$$
If these intervals simultaneously cover their respective score differences, a positive lower endpoint certifies that item $k$ ranks above item $i$, while a negative upper endpoint certifies that it ranks below. We therefore count the items certified above and below each item of interest: $\widetilde n_1(i)=\sum_{k\ne i}\mathbf1\{\widetilde C_L(k,i)>0\}$ and $\widetilde n_2(i)=\sum_{k\ne i}\mathbf1\{\widetilde C_U(k,i)<0\}$. On the simultaneous coverage event, these counts imply $\widetilde n_1(i)+1\le r(i)\le n-\widetilde n_2(i)$ for every $i\in\mathcal I$. The resulting rank confidence sets are
\begin{equation}
\label{eq:rank_ci_pairwise}
\widetilde{\mathcal R}(i)
=\{\widetilde n_1(i)+1,\ldots,n-\widetilde n_2(i)\},
\qquad i\in\mathcal I.
\end{equation}

\noindent\textbf{(ii) One-sided rank bounds and top-$K$ inference.}
To certify that item $i$ lies outside the top-$K$ group, it suffices to obtain a lower confidence bound on $r(i)$ that exceeds $K$. We therefore focus on one-sided inference and construct such a rank bound from simultaneous lower bounds on $\Delta_{ki}^*$. The corresponding one-sided target statistic is
$$
T_{\mathcal I,+}
=\max_{(k,i)\in\mathcal P_{\mathcal I}}
\frac{\widetilde\Delta_{ki}-\Delta_{ki}^*}
{\widetilde\sigma_{ki}}.
$$
Using the same simulated replications as above, its bootstrap analog is
$$
T_{\mathcal I,+}^{(b)}
=\max_{(k,i)\in\mathcal P_{\mathcal I}}
\frac{S_k^{(b)}/g_k^{(b)}-S_i^{(b)}/g_i^{(b)}}
{\widetilde\sigma_{ki}}.
$$
The sign agrees with the leading error in $\widetilde\Delta_{ki}-\Delta_{ki}^*$. Let
$$
\widehat q_{1-\alpha}^{\,\mathrm{lo}}
=\inf\left\{t:
\frac1B\sum_{b=1}^B\mathbf1\{T_{\mathcal I,+}^{(b)}\le t\}
\ge1-\alpha\right\}
$$
be the corresponding one-sided empirical quantile. The simultaneous lower confidence bounds for the score differences are $\widetilde C_L^{\,\mathrm{lo}}(k,i)
=\widetilde\Delta_{ki}
-\widehat q_{1-\alpha}^{\,\mathrm{lo}}\widetilde\sigma_{ki}$ for 
$(k,i)\in\mathcal P_{\mathcal I}$.
On the simultaneous coverage event, $\widetilde C_L^{\,\mathrm{lo}}(k,i)>0$ certifies that item $k$ ranks above item $i$. Thus, define
$\widetilde n_+^{\,\mathrm{lo}}(i)
:=\sum_{k\ne i}\mathbf1
\{\widetilde C_L^{\,\mathrm{lo}}(k,i)>0\}$.
Since at least $\widetilde n_+^{\,\mathrm{lo}}(i)$ items rank above item $i$ on this event, the lower rank bound and its associated confidence set are
\begin{equation}
\label{eq:one_sided_rank_lower_boot}
\widetilde R^{\,\mathrm{lo}}(i)
=1+\widetilde n_+^{\,\mathrm{lo}}(i),
\qquad
\widetilde{\mathcal R}^{\,\mathrm{lo}}(i)
=\{\widetilde R^{\,\mathrm{lo}}(i),\ldots,n\}.
\end{equation}
For $K\in\{1,\ldots,n\}$, consider the hypothesis testing problem
\begin{equation}
H_0:r(i)\le K
\qquad\text{versus}\qquad
H_1:r(i)>K. \label{eq: topk-test}
\end{equation}
We reject $H_0$ whenever $\widetilde R^{\,\mathrm{lo}}(i)>K$. By Theorem~\ref{thm:private_rank_coverage}, the two-sided pairwise intervals and the one-sided lower bounds each have asymptotic simultaneous coverage $1-\alpha$ over $\mathcal P_{\mathcal I}$. Consequently, the induced rank confidence sets satisfy
$$
\liminf_{n\to\infty}
\min\ \left\{\ \mathbb P\{r(i)\in\widetilde{\mathcal R}(i)
\text{ for all }i\in\mathcal I\},\quad 
\mathbb P\{r(i)\in\widetilde{\mathcal R}^{\,\mathrm{lo}}(i)
\text{ for all }i\in\mathcal I\}\ \right\}
\ge1-\alpha.
$$
\section{Theoretical results}\label{sec: theory}

This section establishes theoretical guarantees for the procedures introduced in Section~\ref{sec:private}. We first establish privacy guarantees and estimation-error bounds, then derive the asymptotic expansion underlying inference for individual scores and pairwise differences, and finally establish simultaneous rank coverage and its implication for top-$K$ inference. 

The following proposition provides a privacy guarantee for the proposed estimators and the associated inferential procedures under the edge adjacency relation in Definition~\ref{def:ea}. The inferential outputs use only the private release, public parameters, and independent simulation randomness. They therefore inherit its privacy guarantee by post-processing, without an additional privacy allocation.

\begin{proposition}[Privacy guarantee for the estimator and inferential procedures]
\label{prop:private_est}
If $\lambda > 8/\varepsilon$ and $\gamma > 1/\varepsilon$, then the estimator $\widetilde{\btheta}$ defined in \eqref{eq:dp_est} satisfies $(\varepsilon,0)$-edge differential privacy. Moreover, the score confidence intervals in \eqref{eq:private_ci_score} and \eqref{eq:bc_ci_score}, the pairwise confidence intervals in \eqref{eq:private_ci_compare} and \eqref{eq:bc_ci_pairwise}, and the rank confidence sets in \eqref{eq:rank_ci_pairwise} and \eqref{eq:one_sided_rank_lower_boot} are also $(\varepsilon,0)$-edge differentially private, as are the associated top-$K$ testing decisions.
\end{proposition}

We use the fixed tuning for the statistical guarantees and inference validity,
\begin{equation}
\label{eq:fixed-tuning}
\lambda=c_\lambda/\varepsilon,\qquad
\gamma=c_\gamma/\varepsilon,\qquad c_\lambda>8,\quad c_\gamma>1,
\end{equation}
where the constants are fixed. The constants in the probability and error bounds may depend on $\kappa,c_\lambda,c_\gamma$, but not on $n,p,\varepsilon$. Privacy alone permits larger tuning parameters; the fixed tuning in \eqref{eq:fixed-tuning} supplies the upper bounds needed for statistical accuracy. We take limits along $n\to\infty$ with $p=p_n$ and $\varepsilon=\varepsilon_n$ as specified in each result. Statistical probability statements account for the randomness of the comparison graph, outcomes, and privacy perturbation, together with bootstrap simulation randomness where applicable. 
\begin{theorem}[Finite-sample $\ell_2$ and entrywise error bounds]
\label{thm:bound}
Under Assumption~\ref{assump:bounded_score} and \eqref{eq:fixed-tuning}, assume $p\ge c_0 \log(n)/n$ and $\varepsilon\geq c_0\log(n)^{1.5}/(np)$ for a sufficiently large constant $c_0>0$. Then there exists a constant $C>0$ such that the following inequalities hold with probability at least $1-O(n^{-7})$:
\begin{equation*}
\|\widehat{\btheta}-\btheta^*\|_2^2\le C\left(\frac{1}{p}+\frac{\log(n)^2}{np^2\varepsilon^2}\right)\quad \mbox{ and } \quad \|\widetilde{\btheta}-\btheta^*\|^2_\infty\le  C\left(\frac{\log(n)}{np}+\frac{\log(n)^2}{n^2p^2\varepsilon^2}\right).
\end{equation*}
\end{theorem}
Theorem~\ref{thm:bound} gives the same sampling and privacy rates for entrywise estimation as Theorem~2.1 of \citet{cai2026optimal}, under different tuning choices and conditions on the comparison graph and privacy budget. At a fixed privacy budget $0<\varepsilon<1$, their sufficient lower bound on $p$ is smaller than ours by a factor of $\sqrt{\log n}$. When $np\gg(\log n)^2$, our theorem permits smaller privacy budgets, including values of $\varepsilon$ for which the privacy term dominates the statistical term in the entrywise bound. This regime still includes sparse graphs, for example $p=(\log n)^3/n$. Theorem~\ref{thm:bound} also controls the squared Euclidean error of the centered scores. Its statistical term is of order $1/p$, compared with $\log n/p$ obtained by converting the entrywise guarantee of \citet{cai2026optimal}, while the privacy term has the same order in both bounds.

\citet{cai2026optimal} derive their entrywise bound by tracking leave-one-out gradient iterates, following \citet{chen2019spectral}. We instead use such iterates only to establish preliminary boundedness and then adapt the approach of \citet{chen2022partial} to the perturbed likelihood. For each item, we compare the full fit with the reduced fit obtained by removing that item and its comparisons. Because the reduced fit is independent of the omitted comparisons, we can isolate the individual sampling fluctuation in its score equation. This decomposition yields the entrywise bound. Under the stronger conditions of Theorem~\ref{thm:expansion}, we use a further leave-two-out argument to derive the uniform asymptotic expansion.

\begin{theorem}[Asymptotic expansion for private estimator]
\label{thm:expansion}
Let $\widetilde\btheta$ be the estimator defined in \eqref{eq:dp_est}, let $\bar w=n^{-1}\sum_{i=1}^n w_i$ be the average privacy noise, and write
$$
S_i=\sum_{j\ne i}A_{ij}\{Y_{ij}-F(\theta_i^*-\theta_j^*)\},
\qquad
g_i=\sum_{j\ne i}A_{ij}F'(\theta_i^*-\theta_j^*).
$$
Under Assumption~\ref{assump:bounded_score} and \eqref{eq:fixed-tuning}, assume $np\gg(\log n)^{3/2}$ and $np\varepsilon^2\gg(\log n)^3$. Then, with probability tending to one, uniformly over $i\in[n]$,
$$
\widetilde{\theta}_i-\theta_i^*=(1+r_{1,i})\frac{S_i}{g_i}+r_{2,i}-\frac{\bar w}{\gamma},
$$
where the remainder vectors satisfy $\|\br_1\|_\infty=o_p(1)$ and $\|\br_2\|_\infty=o_p\{(np)^{-1/2}\}$.
\end{theorem}

Centering removes the common shift $-\bar w/\gamma$, which also cancels in pairwise differences. For the centered estimator, the leading term $S_i/g_i$ and the graph condition $np\gg(\log n)^{3/2}$ agree with those of \citet{gao2023uncertainty}. To control the expansion remainder uniformly, we adapt their leave-two-out argument to the perturbed likelihood. Within each leave-one-out fit, removing a second item makes the remaining fitted scores independent of that item's incident comparisons. In contrast, the reduced score equations preserve the ridge penalty and centered privacy perturbation. Under $np\varepsilon^2\gg(\log n)^3$, the resulting ridge bias and centered perturbation are asymptotically negligible. The fixed-coordinate representation \eqref{eq: heuristic expansion} motivates our bias correction and calibration, and the following proposition gives the finite-dimensional normal approximation.

\begin{proposition}[Finite-dimensional Central Limit Theorem]
\label{prop:fd_clt} 
Suppose the conditions of Theorem \ref{thm:expansion} hold. Then, for any prespecified subset $\mathcal{I}\subseteq[n]$ of fixed cardinality,
    $$\rho_i(\btheta^*)^{-1}\{\widetilde{\theta}_i-\operatorname{ave}(\widetilde\btheta)-\theta_i^*\}\stackrel{d}{\to}N(0,1)\text{ and }\rho_{ij}(\btheta^*)^{-1}\{(\widetilde{\theta}_i-\widetilde{\theta}_j)-(\theta_i^*-\theta_j^*)\}\stackrel{d}{\to}N(0,1),$$
    uniformly over $i,j\in\mathcal{I}$ with $i\ne j$ in the pairwise statement, where
    $$\rho_i(\btheta)=\bigg\{\sum_{k\neq i}pF^{\prime}(\theta_i-\theta_k)\bigg\}^{-1/2}\text{ and }\rho_{ij}(\btheta)=\sqrt{\rho_i(\btheta)^2+\rho_j(\btheta)^2}.$$
\end{proposition}

Proposition~\ref{prop:fd_clt} uses oracle standard errors. The next theorem establishes coverage for the feasible Wald and bias-corrected Gaussian--Laplace intervals constructed in Section~\ref{sec: pairwise_ci}.

\begin{theorem}[Plug-in Wald and bias-corrected mixture inference]
\label{thm:plugin_inference}
Suppose that the conditions of Theorem~\ref{thm:expansion} hold. For any fixed $\alpha\in(0,1)$ and prespecified $\mathcal I\subseteq[n]$ with fixed cardinality, the following statements hold.

\noindent\textup{(i) Wald intervals.} The confidence intervals defined in \eqref{eq:private_ci_score} and \eqref{eq:private_ci_compare} satisfy
$$
\max_{i\in\mathcal I}\big|\mathbb P\{\theta_i^*\in\mathrm{CI}_i(1-\alpha)\}-(1-\alpha)\big|\longrightarrow0,
$$
$$
\max_{\substack{i,j\in\mathcal I\\i\ne j}}\big|\mathbb P\{\theta_i^*-\theta_j^*\in\mathrm{CI}_{ij}(1-\alpha)\}-(1-\alpha)\big|\longrightarrow0.
$$

\noindent\textup{(ii) Bias-corrected mixture intervals.} The confidence intervals defined in \eqref{eq:bc_ci_score}--\eqref{eq:bc_ci_pairwise} satisfy
$$
\max_{i\in\mathcal I}\big|\mathbb P\{\theta_i^*\in\mathrm{CI}_i^{(\mathrm{bc})}(1-\alpha)\}-(1-\alpha)\big|\to0,
$$
$$
\max_{\substack{i,j\in\mathcal I\\i\ne j}}\big|\mathbb P\{\theta_i^*-\theta_j^*\in\mathrm{CI}_{ij}^{(\mathrm{bc})}(1-\alpha)\}-(1-\alpha)\big|\to0.
$$
\end{theorem}

Theorem~\ref{thm:plugin_inference} establishes that both the classical plug-in Wald procedure and the bias-corrected Gaussian--Laplace construction attain asymptotic marginal coverage $1-\alpha$ for prespecified scores and pairwise differences. Our coverage guarantees assume negligible asymptotic privacy and regularization effects, and we assess finite-sample performance in Section~\ref{sec:simulation}. Beyond inference for individual pairwise comparisons, inference for even a single item's rank depends on how its score compares with the scores of all $n-1$ opponents. Thus, our bootstrap procedure controls the relevant family of score differences simultaneously. 
\begin{theorem}[Simultaneous rank inference]
\label{thm:private_rank_coverage}
Under Assumption~\ref{assump:bounded_score} and \eqref{eq:fixed-tuning}, assume that $np\gg(\log n)^{5}$ and $np\varepsilon^2\gg(\log n)^3$, and that the number of bootstrap replications satisfies $B\to\infty$. Fix $\alpha\in(0,1)$ and a nonempty prespecified set $\mathcal I\subseteq[n]$, whose cardinality may grow with $n$. Then the following statements hold.

\noindent\textup{(i) Two-sided inference.} The simultaneous intervals with endpoints $\widetilde C_L(k,i)$, $\widetilde C_U(k,i)$ satisfy
$$
\mathbb P\left\{
\Delta_{ki}^*
\in
[\widetilde C_L(k,i),\widetilde C_U(k,i)]
\text{ for all }(k,i)\in\mathcal P_{\mathcal I}
\right\}
\to1-\alpha.
$$
Consequently, the rank confidence sets $\widetilde{\mathcal R}(i)$ satisfy
$$
\liminf_{n\to\infty}
\mathbb P\left\{
r(i)\in\widetilde{\mathcal R}(i)
\text{ for all }i\in\mathcal I
\right\}
\ge1-\alpha.
$$
\noindent\textup{(ii) One-sided inference.} The simultaneous lower bounds $\widetilde C_L^{\,\mathrm{lo}}(k,i)$ satisfy
$$
\mathbb P\left\{
\Delta_{ki}^*
\ge
\widetilde C_L^{\,\mathrm{lo}}(k,i)
\text{ for all }(k,i)\in\mathcal P_{\mathcal I}
\right\}
\to1-\alpha.
$$
Consequently, the one-sided rank confidence sets $\widetilde{\mathcal R}^{\,\mathrm{lo}}(i)$ satisfy
$$
\liminf_{n\to\infty}
\mathbb P\left\{
r(i)\in
\widetilde{\mathcal R}^{\,\mathrm{lo}}(i)
\text{ for all }i\in\mathcal I
\right\}
\ge1-\alpha.
$$
\end{theorem}

The contrast bounds have asymptotic simultaneous coverage $1-\alpha$, while the induced rank coverage can be conservative because simultaneous contrast coverage is sufficient for rank coverage. We achieve this via a Gaussian approximation combined with a bootstrap for the maximum of pairwise score contrasts.
Theorem~\ref{thm:private_rank_coverage} uses the same privacy condition $np\varepsilon^2\gg(\log n)^3$ as Theorem~\ref{thm:expansion}  and strengthens the graph condition to $np\gg(\log n)^5$, which together ensure that both the Gaussian approximation error and privacy-induced perturbation are negligible relative to the anti-concentration scale of the max statistic \citep{chernozhuokov2022improved}. The empirical bootstrap approximation requires $B\to\infty$ without additional privacy composition cost across the bootstrapped samples.

\section{Simulation results}
\label{sec:simulation}

In this section, we evaluate the finite-sample performance of the confidence intervals for scores and pairwise score differences, rank confidence sets, and top-$K$ tests developed in Section~\ref{sec:private}. We generated comparison graphs from $G(n,p)$ and, conditional on each graph, generated one independent Bernoulli comparison on each observed edge under the BT model. We set $K=10$ and constructed the true scores as
\begin{equation}
\label{eq:simulation_scores}
a_i=
\begin{cases}
\displaystyle \frac12+s\frac{K-i}{K-1},&1\le i\le K,\\[5pt]
\displaystyle -\frac12-s\frac{i-K-1}{n-K-1},&K<i\le n,
\end{cases}
\qquad
\theta_i^*=a_i-\frac1n\sum_{j=1}^n a_j.
\end{equation}
This construction gives $r(i)=i$, a boundary gap $\theta_K^*-\theta_{K+1}^*=1$, and a score span of $s$ within each group. Moreover, $\|\btheta^*\|_\infty\le 1+3s/2$, so the choices of $s$ specified below satisfy Assumption~\ref{assump:bounded_score} with $\kappa=4$. The true score vector was held fixed across replications within each configuration. Throughout, we set $\alpha=0.05$ and varied $\varepsilon\in\{\infty,5,2.5,1\}$. For finite $\varepsilon$, we took $c_\gamma=4$ and $c_\lambda=8.1$ in~\eqref{eq:fixed-tuning}. The non-private benchmark, denoted by $\varepsilon=\infty$, used $\widehat{\btheta}_{\mathrm{MLE}}$ with $\gamma=\lambda=0$. We evaluated each configuration over $R=500$ independent Monte Carlo replications. Appendix~D provides additional results on estimation bias and distribution diagnostics of both uncorrected and bias-corrected estimators, and examines the sensitivity of rank inference and top-$K$ placement to the edge sampling probability $p$.

\begin{table}[!htbp]
\centering
\caption{Empirical coverage probabilities and mean lengths of the nominal 95\% Wald and bias-corrected (BC) intervals for the individual score $\theta_K^*$ and score contrast $\theta_K^*-\theta_{K+1}^*$, with $K=10$, $p=0.4$ and $s=0.5$. Each entry reports empirical coverage, with mean interval length in parentheses. Results are based on $R=500$ replications.}
\label{tab:sim_scalar}
\renewcommand{\baselinestretch}{1}\small
\setlength{\tabcolsep}{5pt}
\renewcommand{\arraystretch}{1.1}
\begin{tabular}{ccrrrr}
\toprule
\multirow{2}{*}{$n$} & \multirow{2}{*}{$\varepsilon$} &
\multicolumn{2}{c}{Individual score $\theta_K^*$} &
\multicolumn{2}{c}{Score contrast $\theta_K^*-\theta_{K+1}^*$} \\
\cmidrule(lr){3-4}\cmidrule(lr){5-6}
 & & \multicolumn{1}{c}{Wald} & \multicolumn{1}{c}{BC} &
\multicolumn{1}{c}{Wald} & \multicolumn{1}{c}{BC} \\
\midrule
500 & $\infty$ & 0.968 (0.669) & 0.968 (0.669) & 0.960 (0.876) & 0.960 (0.876) \\
 & 5 & 0.930 (0.665) & 0.948 (0.716) & 0.940 (0.873) & 0.952 (0.929) \\
 & 2.5 & 0.914 (0.662) & 0.958 (0.850) & 0.926 (0.871) & 0.978 (1.092) \\
 & 1 & 0.720 (0.663) & 0.960 (1.551) & 0.708 (0.877) & 0.946 (1.937) \\
\addlinespace[3pt]
700 & $\infty$ & 0.944 (0.565) & 0.944 (0.565) & 0.946 (0.739) & 0.946 (0.739) \\
 & 5 & 0.938 (0.560) & 0.952 (0.589) & 0.944 (0.736) & 0.956 (0.768) \\
 & 2.5 & 0.892 (0.560) & 0.942 (0.675) & 0.916 (0.736) & 0.954 (0.873) \\
 & 1 & 0.740 (0.560) & 0.946 (1.128) & 0.758 (0.738) & 0.952 (1.415) \\
\addlinespace[3pt]
1000 & $\infty$ & 0.948 (0.472) & 0.948 (0.472) & 0.944 (0.617) & 0.944 (0.617) \\
 & 5 & 0.938 (0.471) & 0.944 (0.487) & 0.954 (0.616) & 0.960 (0.634) \\
 & 2.5 & 0.916 (0.470) & 0.944 (0.540) & 0.912 (0.616) & 0.940 (0.695) \\
 & 1 & 0.736 (0.469) & 0.926 (0.822) & 0.764 (0.616) & 0.938 (1.036) \\
\bottomrule
\end{tabular}
\end{table}

\noindent{\bf Individual and pairwise inference.} We first examined confidence intervals for the boundary score $\theta_K^*$ and the boundary contrast $\theta_K^*-\theta_{K+1}^*$, with $n\in\{500,700,1000\}$, $p=0.4$, and $s=0.5$. We compared the Wald intervals in~\eqref{eq:private_ci_score}--\eqref{eq:private_ci_compare} with the bias-corrected (BC) intervals in~\eqref{eq:bc_ci_score}--\eqref{eq:bc_ci_pairwise}. For the BC intervals, we approximated the Gaussian--Laplace quantiles using $M=50{,}000$ Gaussian and Laplace reference draws, generated once independently of the experimental data and reused across replications. We computed the plug-in quantities used to transform these draws separately for each fit. Table~\ref{tab:sim_scalar} reports empirical coverage, with mean interval length in parentheses. Coverage was calculated as the proportion of replications in which the interval contained its target, and length was averaged over the $R$ replications. The procedures coincide in the non-private benchmark.

 At each value of $n$, the Wald interval lengths changed little as $\varepsilon$ decreased, while coverage deteriorated substantially, falling to $0.708$--$0.764$ at $\varepsilon=1$. This pattern was consistent with the Wald calibration accounting only for the leading sampling variation, although regularization bias and privacy perturbation could remain significant in finite samples. The BC procedure combined regularization-bias correction with convolution calibration and produced wider intervals with coverage substantially closer to the nominal level. At fixed $p$, increasing $n$ provided more comparisons per item and shortened the intervals for both procedures. For the BC procedure, the proportional increase in mean length over the non-private benchmark also declined at each fixed privacy level, indicating a smaller relative cost of privacy. We assessed this cost against the non-private benchmark because the shorter private Wald intervals suffered from undercoverage.

\noindent{\bf Rank inference.} We next examined the bootstrap rank confidence sets $\widetilde{\mathcal R}(i)$ in~\eqref{eq:rank_ci_pairwise}, with $n\in\{600,1000,1400\}$, $p=0.6$, and $s=2$. For each item $i\in\{2,5,8\}$, we set $\mathcal I=\{i\}$ and calibrated the comparisons with all other items using $B=7000$ bootstrap draws conditional on the fitted estimator. Coverage therefore refers to each rank, rather than simultaneous coverage of the three displayed ranks. Table~\ref{tab:sim_rank} reports empirical coverage and mean rank endpoints for each item. The bracketed endpoint averages describe both the locations and sizes of the rank confidence sets.

\begin{table}[!htbp]
\centering
\caption{Empirical coverage probabilities and mean lower and upper endpoints of the nominal 95\% rank confidence sets $\widetilde{\mathcal R}(i)$ for $i\in\{2,5,8\}$, with $K=10$, $p=0.6$ and $s=2$. Each target was calibrated separately using $\mathcal I=\{i\}$. For each item, the two columns report empirical coverage and the mean lower and upper endpoints, respectively; the endpoint pair is shown in brackets. Results are based on $R=500$ replications and $B=7000$ bootstrap draws per replication.}
\label{tab:sim_rank}
\renewcommand{\baselinestretch}{1}\small
\setlength{\tabcolsep}{4pt}
\renewcommand{\arraystretch}{1.1}
\begin{tabular}{@{}ccrc@{\hspace{10pt}}rc@{\hspace{10pt}}rc@{}}
\toprule
\multirow{2}{*}{$n$} & \multirow{2}{*}{$\varepsilon$} &
\multicolumn{2}{c}{$i=2$} & \multicolumn{2}{c}{$i=5$} &
\multicolumn{2}{c}{$i=8$} \\
\cmidrule(lr){3-4}\cmidrule(lr){5-6}\cmidrule(lr){7-8}
 & & \multicolumn{1}{c}{Coverage} & Mean endpoints &
\multicolumn{1}{c}{Coverage} & Mean endpoints &
\multicolumn{1}{c}{Coverage} & Mean endpoints \\
\midrule
600 & $\infty$ & 1.000 & [1.00, 6.36] & 1.000 & [1.11, 8.56] & 1.000 & [3.24, 9.96] \\
 & 5 & 1.000 & [1.00, 6.74] & 0.998 & [1.07, 8.75] & 1.000 & [2.88, 9.95] \\
 & 2.5 & 0.996 & [1.00, 7.04] & 0.994 & [1.13, 8.86] & 1.000 & [2.49, 9.95] \\
 & 1 & 0.962 & [1.14, 7.53] & 0.946 & [1.44, 9.14] & 0.970 & [2.56, 12.74] \\
\addlinespace[3pt]
1000 & $\infty$ & 0.998 & [1.00, 5.46] & 1.000 & [1.40, 7.88] & 1.000 & [4.59, 9.84] \\
 & 5 & 1.000 & [1.00, 5.80] & 1.000 & [1.29, 7.97] & 1.000 & [4.38, 9.87] \\
 & 2.5 & 0.998 & [1.02, 6.06] & 0.998 & [1.30, 7.99] & 1.000 & [4.12, 9.85] \\
 & 1 & 0.962 & [1.11, 6.37] & 0.982 & [1.51, 8.28] & 0.986 & [3.67, 9.85] \\
\addlinespace[3pt]
1400 & $\infty$ & 1.000 & [1.00, 4.95] & 1.000 & [1.66, 7.33] & 1.000 & [5.30, 9.73] \\
 & 5 & 0.998 & [1.00, 5.09] & 1.000 & [1.62, 7.49] & 1.000 & [5.16, 9.72] \\
 & 2.5 & 1.000 & [1.01, 5.38] & 1.000 & [1.53, 7.57] & 1.000 & [4.98, 9.75] \\
 & 1 & 0.966 & [1.10, 5.70] & 0.994 & [1.66, 7.82] & 0.994 & [4.44, 9.64] \\
\bottomrule
\end{tabular}
\end{table}

Rank coverage was close to or above the nominal level throughout Table~\ref{tab:sim_rank}, approaching one even in the non-private benchmark. This pattern is consistent with the distinction between score-difference coverage and rank coverage in Theorem~\ref{thm:private_rank_coverage}, as also documented in non-private rank inference \citep{fan2025ranking}. In particular, converting simultaneous confidence intervals for score differences into rank confidence sets can introduce additional conservativeness. The rank confidence sets narrowed as $n$ increased. Stronger privacy generally enlarged the sets, although at $n=1400$ the mean cardinality under $\varepsilon=1$ exceeded the non-private benchmark by less than one rank for each of the three displayed items. Even without privacy, the sets still spanned several rank positions on average, indicating that uncertainty about the items' exact ranks remained.

\begin{figure}[!htbp]
\centering
\includegraphics[width=0.8\textwidth]{ 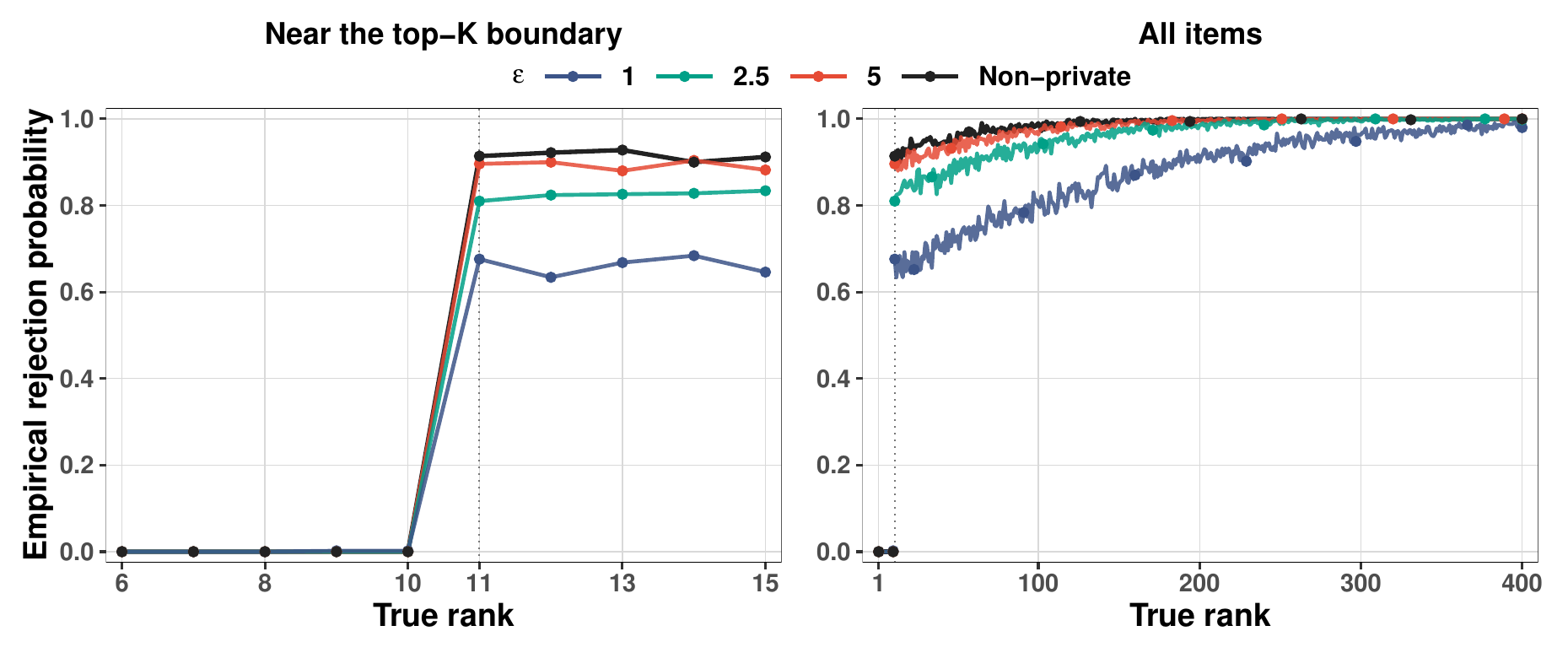}
\caption{Empirical rejection probabilities of the individual top-$K$ placement tests over $R=500$ replications, with $n=400$, $p=0.6$, $K=10$, $s=0.5$, and $B=7000$ bootstrap draws per replication. The left panel shows ranks $6$--$15$ and the right panel shows all items. Colored curves give results for finite $\varepsilon$; the black curve gives the non-private benchmark. Vertical dotted lines mark the first rank outside the top-$K$ group. Each item was calibrated separately.}
\label{fig:sim_placement}
\end{figure}

\noindent{\bf Top-K inference.} Finally, we evaluated the proposed tests for the individual top-$K$ problem in~\eqref{eq: topk-test}, with $n=400$, $p=0.6$, and $s=0.5$. We used the one-sided rank bound $\widetilde R^{\,\mathrm{lo}}(i)$ in~\eqref{eq:one_sided_rank_lower_boot} and rejected when $\widetilde R^{\,\mathrm{lo}}(i)>K$. Each item was calibrated separately using $\mathcal I=\{i\}$ and $B=7000$ bootstrap draws. Figure~\ref{fig:sim_placement} shows the empirical rejection probabilities near the top-$K$ boundary and over the full range of ranks, calculated as the proportion of rejections across the $R$ replications. Rejection probability measures type I error for the items with $r(i)\le K$ and power for those with $r(i)>K$.

Empirical rejection probabilities for the ten null items were below the nominal level at every privacy budget, consistent with the asymptotic type I error control implied by Theorem~\ref{thm:private_rank_coverage}. Under moderate privacy, power was close to the non-private benchmark over much of the rank range. More pronounced losses occurred under stronger privacy, particularly for items near the top-$K$ boundary, whereas power approached one farther from the boundary even at the strongest privacy level considered. In this configuration, the principal cost of stronger privacy was therefore a reduced ability to identify and exclude items near the boundary, while erroneous exclusions of true top-$K$ items remained rare.

\section{Real data analysis}
\label{sec:real_data}

Creative-writing platforms face the practical problem of screening a growing pool of language models before committing resources to detailed editorial evaluation. For short-form drama, such as that distributed through platforms including {\sl Hongguo} and {\sl ReelShort}, language models could assist by drafting scenes for subsequent editorial development. We consider a creative-writing task that asks for a dramatic scene of 500--700 words based on \emph{Hamlet} under the premise that Claudius neither killed King Hamlet nor arranged his death. The familiar characters and specified premise provide a common narrative foundation for comparing models' ability to develop coherent dramatic conflict, characterization, and dialogue. Pairwise comparisons of the generated scenes provide evidence for selecting models suited to this task. An initial screening stage may exclude individual models whose performance falls outside a target top-$K$ group, while limiting the risk of discarding a genuinely top-$K$ candidate. 
When screening reports are shared with model providers or publishing partners, the underlying editorial or reader assessments may need to remain confidential to protect individual comparison records. This motivates top-$K$ inference that accounts for both sampling uncertainty and privacy perturbation under edge differential privacy.

Using the October 15, 2026 snapshot of the Arena text leaderboard \citep{chiang2024chatbot}, we analyzed $n=99$ models with accessible application programming interfaces (APIs) from the 402 listed models. The original pool contained $100$ models, but one lost every comparison and was excluded; the privacy guarantee applies to each fit on the $n=99$ models. We constructed a complete set of pairwise comparisons among these models. For each pair $(i,j)$, both models generated fresh responses independently of those used for other pairs. The responses were then evaluated by GPT-6.1 Sol using the task prompt and task-specific criteria, with model identities replaced by anonymous labels. We recorded $Y_{ij}=1$ if model $i$ was preferred over model $j$, and $Y_{ij}=0$ otherwise. Here we used an automated judge for simplicity, and the proposed procedure also accommodates human pairwise judgments and does not require the prompt or evaluation criteria to be public. 

We then illustrate the proposed inference procedures using the generated comparison data under a working BT model. With the observed judgments fixed, we sampled $R=500$ graphs from $G(n,p)$ at each $p\in\{0.6,0.8\}$ and fitted the model on each graph at $\varepsilon\in\{\infty,5,2.5,1\}$, using the tuning constants in Section~\ref{sec:simulation}. Each finite $\varepsilon$ specifies the privacy budget for one fit, with fresh perturbations across repetitions. For $K=10$ and $\alpha=0.05$, we constructed the one-sided lower rank bound $\widetilde R^{\,\mathrm{lo}}(i)$ in~\eqref{eq:one_sided_rank_lower_boot} and excluded model $i$ when $\widetilde R^{\,\mathrm{lo}}(i)>K$. Each model was calibrated separately using $\mathcal I=\{i\}$ and $B=7000$ bootstrap draws. These repetitions assess the sensitivity of individual rank inference to comparison subsampling and privacy perturbation, conditional on the observed judgments. Table~\ref{tab:real_hamlet} presents four illustrative models spanning different positions in the non-private ordering estimated from all observed comparisons. The non-private summaries include only repetitions yielding finite MLEs, whereas the private summaries use all $500$ repetitions.

\begin{table}[!htbp]
\centering
\caption{One-sided lower rank bounds $\widetilde R^{\,\mathrm{lo}}(i)$ and top-$10$ exclusion rates for four selected models at $\alpha=0.05$. For each model and setting, we report the median and interquartile range $[Q_{25},Q_{75}]$ of $\widetilde  R^{\,\mathrm{lo}}(i)$ across repetitions. Exclusion (\%) is the percentage of repetitions in which $\widetilde  R^{\,\mathrm{lo}}(i)>10$. Results for each finite $\varepsilon$ use all $500$ fits, whereas the non-private benchmark ($\varepsilon=\infty$) includes only the $233$ and $382$ fits yielding finite MLEs at $p=0.6$ and $0.8$, respectively.}
\label{tab:real_hamlet}
\renewcommand{\baselinestretch}{1}\small
\setlength{\tabcolsep}{3pt}
\renewcommand{\arraystretch}{1.1}
\begin{tabular}{@{}lccr@{\hspace{18pt}}cr@{}}
\toprule
\multirow{2}{*}{Model} & \multirow{2}{*}{$\varepsilon$} &
\multicolumn{2}{c}{$p=0.6$} & \multicolumn{2}{c}{$p=0.8$} \\
\cmidrule(lr){3-4}\cmidrule(lr){5-6}
 & & Median [$Q_{25},Q_{75}$] & \multicolumn{1}{c}{Exclusion (\%)} &
Median [$Q_{25},Q_{75}$] & \multicolumn{1}{c}{Exclusion (\%)} \\
\midrule
\multirow[t]{4}{*}{\parbox[t]{94pt}{\raggedright GPT-5.5}} & $\infty$ & \makebox[12pt][r]{1}\,\makebox[40pt][l]{[1, 1]} & 0.0 & \makebox[12pt][r]{1}\,\makebox[40pt][l]{[1, 1]} & 0.0 \\
 & 5 & \makebox[12pt][r]{1}\,\makebox[40pt][l]{[1, 1]} & 0.0 & \makebox[12pt][r]{1}\,\makebox[40pt][l]{[1, 1]} & 0.0 \\
 & 2.5 & \makebox[12pt][r]{1}\,\makebox[40pt][l]{[1, 1]} & 0.0 & \makebox[12pt][r]{1}\,\makebox[40pt][l]{[1, 1]} & 0.0 \\
 & 1 & \makebox[12pt][r]{1}\,\makebox[40pt][l]{[1, 1]} & 4.6 & \makebox[12pt][r]{1}\,\makebox[40pt][l]{[1, 1]} & 3.6 \\
\addlinespace[5pt]
\multirow[t]{4}{*}{\parbox[t]{94pt}{\raggedright Claude Sonnet 4.6}} & $\infty$ & \makebox[12pt][r]{1}\,\makebox[40pt][l]{[1, 1]} & 0.0 & \makebox[12pt][r]{1}\,\makebox[40pt][l]{[1, 2]} & 0.0 \\
 & 5 & \makebox[12pt][r]{1}\,\makebox[40pt][l]{[1, 1]} & 0.0 & \makebox[12pt][r]{1}\,\makebox[40pt][l]{[1, 1]} & 0.0 \\
 & 2.5 & \makebox[12pt][r]{1}\,\makebox[40pt][l]{[1, 1]} & 0.8 & \makebox[12pt][r]{1}\,\makebox[40pt][l]{[1, 1]} & 0.4 \\
 & 1 & \makebox[12pt][r]{1}\,\makebox[40pt][l]{[1, 3]} & 4.6 & \makebox[12pt][r]{1}\,\makebox[40pt][l]{[1, 3]} & 3.8 \\
\addlinespace[5pt]
\multirow[t]{4}{*}{\parbox[t]{94pt}{\raggedright Kimi-K2}} & $\infty$ & \makebox[12pt][r]{6}\,\makebox[40pt][l]{[5, 7]} & 2.1 & \makebox[12pt][r]{7}\,\makebox[40pt][l]{[7, 8]} & 2.4 \\
 & 5 & \makebox[12pt][r]{2}\,\makebox[40pt][l]{[1, 4]} & 0.8 & \makebox[12pt][r]{4}\,\makebox[40pt][l]{[3, 6]} & 2.2 \\
 & 2.5 & \makebox[12pt][r]{2}\,\makebox[40pt][l]{[1, 3]} & 3.4 & \makebox[12pt][r]{3}\,\makebox[40pt][l]{[2, 5]} & 3.8 \\
 & 1 & \makebox[12pt][r]{3}\,\makebox[40pt][l]{[1, 6]} & 14.0 & \makebox[12pt][r]{4}\,\makebox[40pt][l]{[2, 7]} & 15.4 \\
\addlinespace[5pt]
\multirow[t]{4}{*}{\parbox[t]{94pt}{\raggedright DeepSeek-V3.1}} & $\infty$ & \makebox[12pt][r]{21}\,\makebox[40pt][l]{[19, 23]} & 100.0 & \makebox[12pt][r]{23}\,\makebox[40pt][l]{[22, 24]} & 100.0 \\
 & 5 & \makebox[12pt][r]{14}\,\makebox[40pt][l]{[11, 17]} & 80.6 & \makebox[12pt][r]{19}\,\makebox[40pt][l]{[16, 20]} & 98.4 \\
 & 2.5 & \makebox[12pt][r]{11}\,\makebox[40pt][l]{[8, 16]} & 54.8 & \makebox[12pt][r]{16}\,\makebox[40pt][l]{[13, 20]} & 86.8 \\
 & 1 & \makebox[12pt][r]{9}\,\makebox[40pt][l]{[4, 17]} & 44.0 & \makebox[12pt][r]{13}\,\makebox[40pt][l]{[7, 19]} & 60.4 \\
\bottomrule
\end{tabular}
\end{table}

At each fixed $p$, GPT-5.5 and Claude Sonnet 4.6 were occasionally excluded under stronger privacy, although their median lower rank bounds remained at $1$. For Kimi-K2, at $p=0.8$, the exclusion frequency increased from $2.2\%$ at $\varepsilon=5$ to $15.4\%$ at $\varepsilon=1$, while the median bound remained at $4$. Thus, the increase in exclusions reflected changes in the upper tail rather than a shift in the typical bound beyond the top-$10$ threshold. For DeepSeek-V3.1, whose bounds mostly exceeded $10$ at $\varepsilon=5$, stronger privacy reduced the median and widened the interquartile range across the threshold, leading to more non-exclusions. These patterns are consistent with the combined effects of stronger regularization and larger privacy perturbations in the private score estimators. Increasing $p$ partly offset this loss of exclusion evidence for DeepSeek-V3.1: at $\varepsilon=1$, its exclusion frequency rose from $44.0\%$ to $60.4\%$ as $p$ increased from $0.6$ to $0.8$. This suggests that additional comparisons strengthen the information about score differences relative to the regularization and privacy perturbations at a fixed privacy budget. Appendix~E.2 presents a complementary analysis of a story-beginning task from MT-Bench \citep{zheng2023judging} that examines the same four illustrative models. The comparison illustrates how the evidence for individual top-$K$ exclusion can vary across tasks, highlighting the importance of choosing evaluation tasks and criteria suited to the intended application.

\FloatBarrier

\section{Discussion}\label{sec:discussions}

Motivated by the use of pairwise comparisons to evaluate LLMs, this paper develops a framework for ranking inference that protects individual preference records. Under the BT model, we establish estimation guarantees and asymptotically valid procedures for score and rank inference. Simulation studies demonstrate the benefits of the proposed bias correction and calibration for finite-sample coverage. The real-data application illustrates how the proposed procedures can support selecting LLMs for creative writing.

Several directions are worth further investigation. First, future work could extend current inference to stronger privacy regimes in which regularization bias can dominate statistical sampling variation. Establishing coverage guarantees for the proposed bias-aware confidence intervals in these regimes would require sharper distributional approximations and tighter control of bias-correction error. Second, heterogeneous comparison designs and covariate-dependent comparison models would accommodate more flexible evaluation settings. Incorporating prompt characteristics could capture variation in relative model performance across tasks, while online extensions would require accounting for adaptive comparison selection and privacy expenditure across successive releases. Third, extending privacy protection from individual comparison records to all judgments contributed by a participant would require user-level differential privacy. This extension would involve controlling the influence of multiple contributions, including repeated evaluations of the same pair, and accounting for dependence among judgments in the inferential analysis. We leave these to future work.

\bibliographystyle{apalike}
\bibliography{ ref}

\end{document}